\documentclass[graybox]{svmult}

\usepackage{type1cm}        
\usepackage{booktabs}
\usepackage{makeidx}         
\usepackage{graphicx}        
\usepackage{multicol}        
\usepackage[bottom]{footmisc}

\usepackage{newtxtext}       %
\usepackage[varvw]{newtxmath}       

\usepackage{algorithm}
\usepackage[noend]{algpseudocode}
\usepackage{wrapfig}

\usepackage{cleveref}
\usepackage{todonotes}

\makeindex             

\DeclareMathOperator*{\argmin}{arg\,min}

\begin{document}

\title*{Two-level deep domain decomposition method}
\author{Victorita Dolean\orcidID{0000-0002-5885-1903} \and Serge Gratton\orcidID{0000-1111-2222-3333} \and Alexander Heinlein\orcidID{0000-0003-1578-8104} \and Valentin Mercier\orcidID{0000-1111-2222-3333}}
\institute{Victorita Dolean \at Eindhoven University of Technology, Department of Mathematics and Computer Science, PO Box 513, 5600 MB Eindhoven, the Netherlands \email{v.dolean.maini@tue.nl}
\and Serge Gratton \at Universit\'e de Toulouse, INP-ENSEEIHT, IRIT, ANITI, Toulouse,France. \email{serge.gratton@toulouse-inp.fr}
\and Alexander Heinlein \at Delft University of Technology, Delft Institute of Applied Mathematics, Mekelweg 4, 2628 CD Delft, Netherlands \email{a.heinlein@tudelft.nl}
\and Valentin Mercier \at Universit\'e de Toulouse, ANITI, CERFACS, IRIT, Toulouse, and BRLi, France.  \email{valentin.mercier@toulouse-inp.fr}
}
%
%
\maketitle

\abstract*{This study presents a two-level Deep Domain Decomposition Method (Deep-DDM) augmented with a coarse-level network for solving boundary value problems using physics-informed neural networks (PINNs). The addition of the coarse level network improves scalability and convergence rates compared to the single level method. Tested on a Poisson equation with Dirichlet boundary conditions, the two-level deep DDM demonstrates superior performance, maintaining efficient convergence regardless of the number of subdomains. This advance provides a more scalable and effective approach to solving complex partial differential equations with machine learning.}


\section{Introduction}
The successful application of machine learning in image and language processing 
has extended its reach to solving physical equations. 
One example for the combination of scientific computing and machine learning is the field of physics-informed machine learning, with physics-informed neural networks (PINNs~\cite{raissi2017physics}) becoming its most prominent example. The idea of this method is to integrate the partial differential equation directly into the loss function for training a neural network to approximate the solution of a boundary value problem (BVP). Therefore, let us consider the generic BVP:
\begin{align*}
    \mathcal{N}[u](x) &= f(x) \; \forall x \in \Omega \\
    \mathcal{B}[u](x) &= g(x) \; \forall x \in \partial \Omega
\end{align*}
with $\Omega \in \mathbf{R}^d$ and $\partial \Omega$ its boundary, $\mathcal{N}$ a differential and $\mathcal{B}$ a boundary operator. 

The approximation of this problem using a neural network $u_\theta$ parametrized by weights and biases gathered in $\theta$ can be found by the following optimization problem: 
$$
	\theta^* = \argmin_\theta \mathcal{M}(\theta),
$$ 
where
\begin{equation}
    \label{loss}
    \mathcal{M}(\theta) = \frac{\lambda_\Omega}{N_\Omega}\sum_{i=1}^{N_\Omega}(\mathcal{N}[u_\theta](x_i)-f(x_i))^2 + \frac{\lambda_{\partial\Omega}}{N_{\partial\Omega}}\sum_{i=1}^{N_{\partial\Omega}}(\mathcal{B}[u_\theta](\hat x_i)-g(\hat x_i))^2
\end{equation}
Here, $\{x_i\}_{i=1}^{i=N_\Omega}$ and $\{\hat x_i\}_{j=1}^{j=N_{\partial\Omega}}$ are sets of collocation points sampled in $\Omega$ respectively on $\partial \Omega$. The back-propagation algorithm allows for both the evaluation of the residual of the partial differential equation (PDE) and the optimization of the loss function~\cref{loss} with respect to the network parameters $\theta$.

While domain decomposition methods (DDMs) are well-established solvers for PDEs using classical discretizations, the use of neural network-based discretizations, in particular, PINNs has been explored more recently.
This concept has been explored in various studies like for example in~\cite{shukla_parallel_2021}, where the authors discuss non-overlapping DDMs for parallel training. Extensions to Schwarz methods for Deep Ritz networks, which integrate the variational form into the loss function, are presented in~\cite{li_d3m_2020}. The coupling in the finite basis PINNs (FBPINNs) approach in~\cite{moseley2021finite} differs from the aforementioned approaches; the authors introduce an overlapping domain decomposition, and the coupling is performed via a corresponding partition of unity scaling and hard enforcement of boundary conditions. For a broader overview over the combination of domain decomposition and machine learning methods, we refer to the review~\cite{klawonn2023machine}. 

Our focus is on employing PINNs as the subdomain solver in a classical Schwarz approach iteration; this is known as the Deep Domain Decomposition Method (Deep-DDM) introduced in~\cite{li_deep_2020}. Given that the subdomain problems can be solved sufficiently accurately, the convergence properties are the same as for the classical Schwarz iteration. As a result, the method is not numerically scalable when increasing the number of subdomains.
In this paper, we will incorporate a coarse level, in order to retain numerical scalability. A related two-level Schwarz approach, which uses a different coupling between the two levels of a classical Schwarz iteration has been presented in~\cite{jang_partitioned_2023}.
Other related multilevel approaches are, for instance, 
multilevel FBPINNs~\cite{dolean2023finite, dolean:MDD:2024} and multilevel optimization methods using frequency-aware networks%
~\cite{Gratton2024}. 


\section{The Deep Domain Decomposition Method}

\begin{figure}
    \centering
    \includegraphics[width=0.55\linewidth]{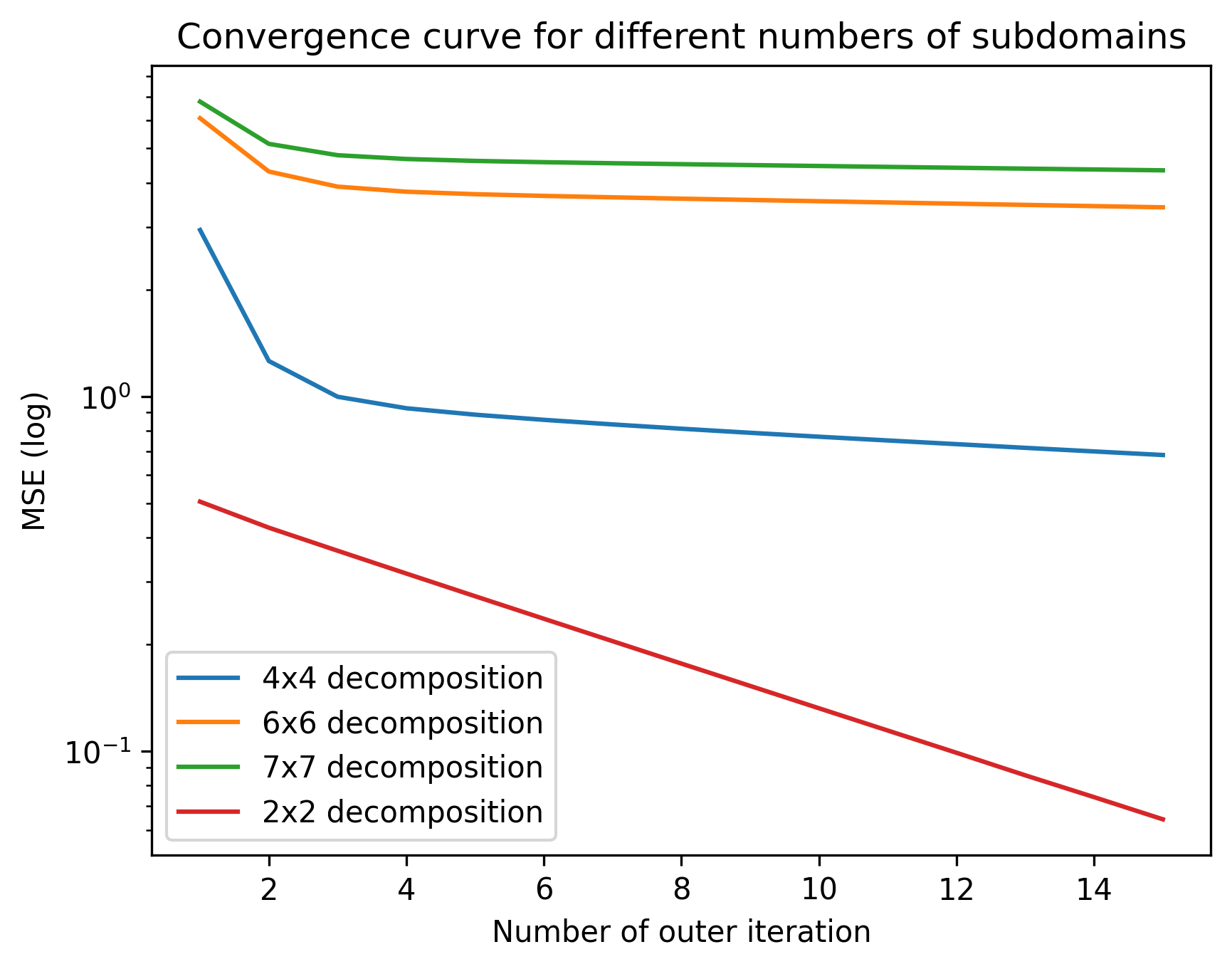} \hfill
    \includegraphics[width=0.35\textwidth]{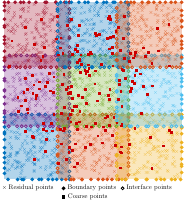}
    \caption{ (1) Strong scalability test on the Deep-DDM method (2) Sampling points for the two-level Deep-DDM method}
    \label{fig:strong-scal}
\end{figure}

Let the computational domain $\Omega$ be decomposed into $S$ overlapping subdomains $\Omega_1,\ldots,\Omega_S$.
Then, we consider the classical Schwarz iteration~\cite{schwarz_ueber_1870}, which involves an independent problem on each subdomain $\Omega_s$: find $u_s$
\begin{equation}
\label{local_problem}
    \left\{
    \begin{aligned}
    &\mathcal{N}(u_s)= f &&  \text{in } \Omega_s,  \\
    &\mathcal{B}(u_s) = g && \text{on }  \partial \Omega_s \backslash \Gamma_s, \\
    &\mathcal{D}(u_s) = \mathcal{D}(u_r) && \text{on }  \Gamma_s,
    \end{aligned}
  \right.
\end{equation}
where $\mathcal{D}$ is an operator for the transmission conditions (e.g., Dirichlet, Neumann, or Robin) on the subdomain boundary, and $\Gamma_s$ is the interface between the subdomain $\Omega_s$ and the neighboring subdomains $\Omega_r$, with $\Omega_r \cap \Omega_s \neq \emptyset$. For convenience, we define the multiple overlapping subdomains as a single subdomain $\Omega_r$ with a corresponding solution network $u_r$.
In order to train a PINN model to solve~\cref{local_problem}, we incorporate an additional term accounting for the transmission conditions into the loss function~\cref{loss}. In particular, the loss term is computed by sampling points on the interface denoted $\{\tilde x_i\}_{i=1}^{N_{\Gamma}} \subset \partial\Omega_r$. A visualization of all sampling points is shown in~\cref{fig:strong-scal} (2). 
For the transmission to $\Omega_r$, with $\Omega_r \cap \Omega_s \neq \emptyset$, we incorporate the loss term: 
\begin{equation}
\label{equ:interface}
\begin{aligned}
    \mathcal{M}_\Gamma(\theta) &= \frac{1}{N_{\Gamma}}\sum_{i=1}^{N_{\Gamma}}|\mathcal{D}({u_s(\tilde x_k)})- W_i|^2
\end{aligned}
\end{equation}
In this term, we minimize the difference between the trained network $u_s$ and the neighboring networks $u_r$ on the interface with respect to the transfer operator $\mathcal{D}$. Here, 
\begin{equation} \label{trans}
	W_i = \mathcal{D}({u_r(\tilde x_k)})
\end{equation}
is defined based on the network $u_r$ from the previous outer Schwarz iteration.
Once all the local subnetworks have been trained up to a certain stopping criterion (e.g., number of iterations or tolerance), the interface values $W_i$ are being updated, and we proceed to the next outer iteration.

The one-level Deep DDM algorithm, without the \textcolor{red}{red} parts, is shown in ~\cref{alg:two-level-Deep-DDM}. Initially, two stopping criteria were used to assess convergence: differences in the network solution between subsequent iterations, both in the interior and at the boundary. In addition, a loss criterion variation was used to terminate the training of each network, alongside a maximum epoch limit. This method made comparisons between runs difficult as the number of epochs varied. Here we simplify by using a fixed number of epochs per training and fixed outer iterations. For further details on the original criteria, refer to~\cite{li_deep_2020}.


\begin{center}
	\setlength{\textfloatsep}{0pt}
	\setlength{\floatsep}{0pt}
	\setlength{\intextsep}{0pt}
	\begin{algorithm}[htp]
		\caption{Two-level DeepDDM; \textcolor{red}{coarse level}}\label{alg:two-level-Deep-DDM}
		\begin{algorithmic}[1]
			\State Sampling the fine \textcolor{red}{and the coarse} collocation points
			\State Initialization of the network parameters $\theta_s^0$ \textcolor{red}{and $\theta_c$}
			\State Initialization of interface values $W=[W_1,...,W_s]$
			\State \textcolor{red}{Initialization of weights $\lambda_f$ and $\lambda_c$}
			\While{Iteration limits not reached}
			\State Local network training
                \State \textcolor{red}{Coarse network training}
			\State \textcolor{red}{Compute $\sum_{s=1}^S E_s(\chi_s u_s(x_{i,coarse})$ for each coarse points}
			\State Update of $W_k$ values at interfaces with \cref{equ:interface} \textcolor{red}{or \cref{equ:newwk}}
			\State \textcolor{red}{Update $\lambda_f$ and $\lambda_c$}
			\EndWhile
		\end{algorithmic}
	\end{algorithm}
\end{center}

As mentioned in the introduction, the convergence of the one-level method does not scale with the number subdomains.
We consider 
two types of scalability:
\begin{itemize}
    \item \emph{Strong scalability:} Strong scalability is defined as how the solution time varies with the number of cores for a fixed total problem size. Ideally, the elapsed time is inversely proportional to the number of processing units. 
    \item \emph{Weak scalability:} Weak scalability is defined as how the solution time varies with the number of cores for a fixed problem size per core. Ideally, the elapsed time is constant for a fixed ratio between the size of the problem and the number of processing units. 
\end{itemize}

Since these terms are not clearly defined for of deep learning, we will assume that the size of our problem is the number of points sampled to solve the problem, and the number of processing units is the number of PINNs models used to solve the problem (and therefore the number of subdomains).  

We consider a Poisson equation with Dirichlet boundary conditions:
\begin{equation*}
\label{probleme_poisson}
  \left\{
  \begin{aligned}
      &\Delta u = r(x) && \text{in }\Omega = [0,1]\times[0,1], \\   
      &u = g(x) && \text{on } \partial\Omega. \\
  \end{aligned}
  \right.
\end{equation*}

We choose $r$ and $g$ such that the exact solution is $ u(x) = sin(\omega_1\pi x_1)sin(\omega_1\pi x_2) + sin(\omega_2\pi x_1)sin(\omega_2\pi x_2)$, in this paper we will perform tests for several values of $\omega_1$ and $\omega_2$. To investigate the scalability of the one-level method we choose $\omega_1=\omega_2=1$ and we perform 1500 epochs per training (other settings are the same as in \ref{sec:num}). Here, we test the strong scalability on regular rectangle domain decomposition. A regular rectangle domain decomposition with overlap divides $\Omega = [L_{x_0}, L_{x_1}] \times [L_{y_0}, L_{y_1}]$ into $N_x \times N_y$ subdomains, each overlapping by $\alpha_x \Delta x$ and $\alpha_y \Delta y$. The total number of points sampled remains the same in each experiment, corresponding to our problem size, while we increase the number of subdomains and thus the number of processing units. In~\cref{fig:strong-scal} (1), the mean squared error over outer Schwarz iterations for several numbers of subdomains are plotted. We observe that the convergence deteriorates when increasing the number of subdomains. Therefore, in the next section, we will introduce a coarse level for the Deep-DDM, which facilitates fast global transport of information, to retain scalability. 

\section{Extension via a Coarse Network}
Our coarse level corresponds to training a neural network acting on the entire domain, which we will denote as the \emph{coarse network}. The convergence of the two-level method will then depend on the coarse network solution as well as the exchange of information with the local networks on the first level of the method. 
In particular, as the coarse network, we train a classical PINN model on the global domain and add an additional loss term incorporating the local subdomain networks. Conversely, after training the coarse network, we will incorporate the coarse network into the loss function for the training of the local networks.

\noindent \textbf{Extension operators and a partition of unity~\cite{dolean_introduction_2015}:} Let us define an extension operator 
$$
	E_s(w_s) = \left\lbrace \begin{array}{ll}
		w_s & \text{ in } \Omega_s \\
		0 & \text{ otherwise.} 
	\end{array} \right.
$$
Here, $w_s$ is a function defined on $\Omega_s$.
Moreover, we define partition of unity functions $\chi_s$ with $\chi_s\geq 0$, ${\rm supp} (\chi_s) \subset \overline{\Omega_s}$, $\chi_s(x)=0$ for $x \in \partial \Omega_s \backslash \partial \Omega$, and
$$ 
	w = \sum_{s=1}^S E_s(\chi_s w|_{\Omega_s})
$$
for any function $w$ defined on $\Omega$. We then add the term
$$
    \mathcal{M}_{fine}(\theta_c) = \frac{1}{N_{\Omega_c}}\sum_{k=1}^{N_{\Omega_c}} \big| u_{c}(x_{k}^c) - \sum_{s=1}^S E_k(\chi_s u_s(x_{k}^c)) \big|^2,
$$

to the loss function~\cref{loss} for the training of the coarse network. Here, $\{x_k^c\}_{k=1}^{k=N_{\Omega_c}}$ are the sampling points for the coarse network $\Omega$ related loss (example in ~\cref{fig:strong-scal} (2)), and $u_c$ and $\theta_c$ are the coarse network and its network parameters, respectively. This term which transfer information from the fine solution to the coarse network is weighted with $\lambda_f$. As we aim to train the coarse network concurrently with the fine networks, making our algorithm fully parallelizable, we use the \( u_s \) from the previous outer iteration. This information is unavailable for the first iteration, so in this case, we set \( \lambda_f \) to 0.


In order to transfer information from the coarse to the local networks, we introduce a loss term into the $M_\Gamma$ loss of the local networks via the interface term~\cref{trans}:
\begin{equation} \label{equ:newwk}
	W_i = \lambda_c\mathcal{D}({u_r(\tilde x_i)}) + (1-\lambda_c)\mathcal{D}({u_c(\tilde x_i)}),
\end{equation}
where $\lambda_c \in [0,1]$ is a weight balancing the impact of the coarse network on the local networks.

Note that, since the accuracy of the local networks is often poor in the first iterations, we adjust the corresponding weight in the loss function $\lambda_f$ during the Schwarz iteration. 
%
The final two-level algorithms is given in~\cref{alg:two-level-Deep-DDM}, including the \textcolor{red}{red} parts.

\section{Numerical Results}
\label{sec:num}

In this section, we will conduct tests on the previously defined Poisson's problem using the following settings.
The collocation points are sampled using Latin hypercube sampling in $\Omega$ and on $\partial\Omega$ and $\Gamma$. For our strong scaling tests, we fix $N_\Omega = 30\,000$ and $N_{\partial\Gamma}=N_{\Gamma}=16\,000$ for the whole problem while increasing the number of subdomains. 
All networks are trained using the Adam optimizer with an initial learning rate of $2\times 10^{-4}$  and an exponential decay of 0.999 every 100 epochs. Each neural network is composed of two hidden layers with 30 neurons. The overlap is set to $30 \%$ of the subdomain larger side.
In each Schwarz iteration, each local and coarse network is trained for 2\,500 epochs. The weight controlling the impact of the coarse network on the local networks is set to $\lambda_c = 1\times0.9^{I}$ where $I$ is the index of the Schwarz iteration. This increases the impact of the coarse network during the course of the outer Schwarz iteration.
The coefficient $\lambda_f$ controlling the impact of the fine networks on the coarse network is set to a fixed value of $0.5$. All weights $\lambda_*$ have been optimized using a rough grid search to obtain good performance. 

Our implementation uses TensorFlow2 (version 24.02) and runs numerical experiments on a single Nvidia A100-80 GPU. Both the fine and coarse network training processes run concurrently on the GPU using multiprocessing. Although this setup is suboptimal due to the unaddressed GPU load and the complex interaction between TensorFlow and multiprocessing, it allows us to compare the wall time between one-level and two-level methods. The results presented are the median of three independent training runs, each initialized with different seeds.





We focus on investigating the impact of our two-level domain decomposition on the \emph{spectral bias} or \emph{f-principle}~\cite{Xu20,Wang_2021} of neural networks; this refers to
the observation low-frequency components of the target functions are learned much faster than the high-frequency components.
The Deep-DDM method tackles this issue by splitting the global problem into smaller subproblems, allowing for better approximation of high frequency components, while the coarse network is supposed to learn the low frequency components.

In order to investigate this, we consider two different pairs of coefficients $(\omega_1,\omega_2) \in \left\lbrace (1,3); (1,6) \right\rbrace$.
\begin{figure}[t]
\centering
\includegraphics[width=0.5\textwidth]{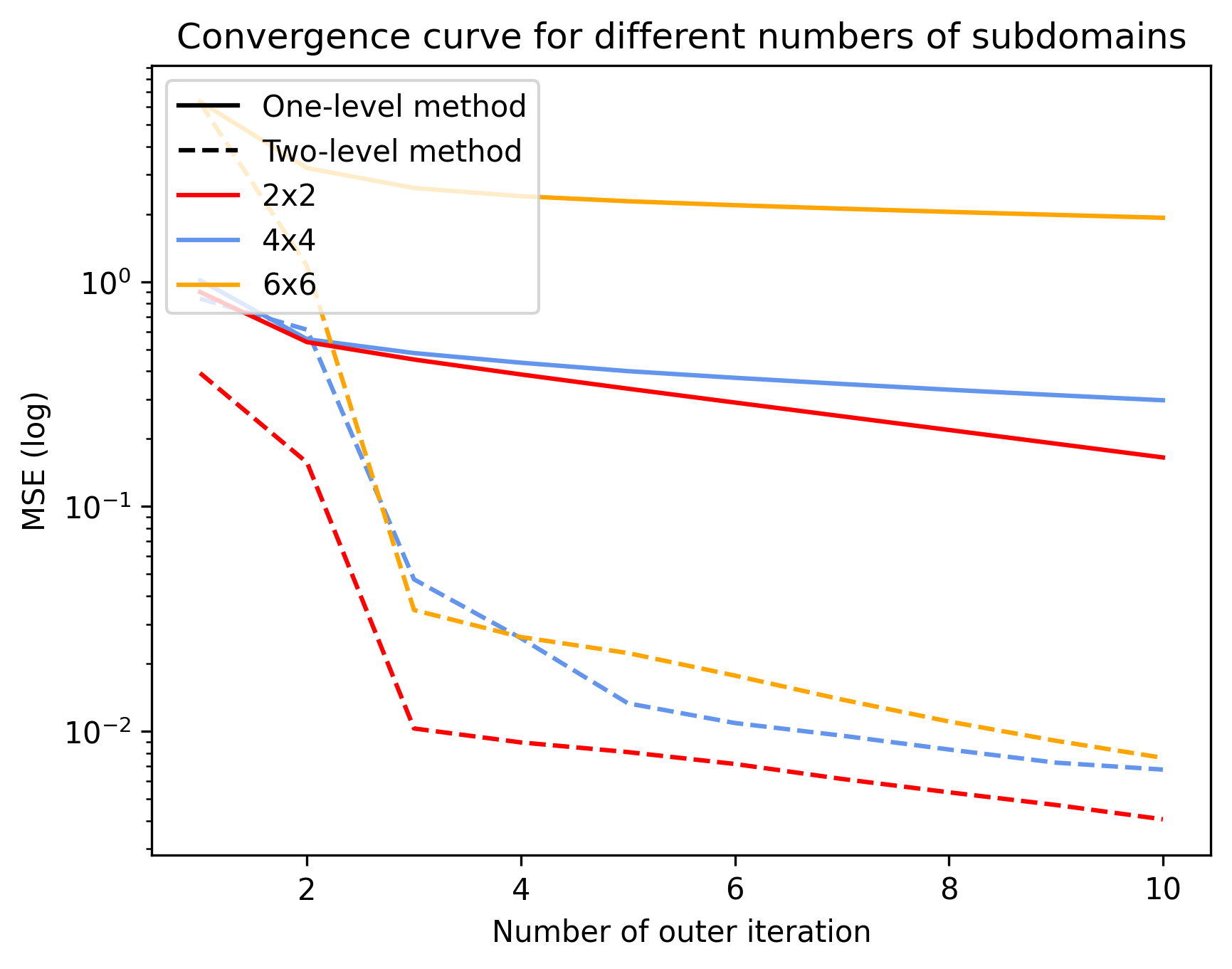}\hfill
\includegraphics[width=.5\textwidth]{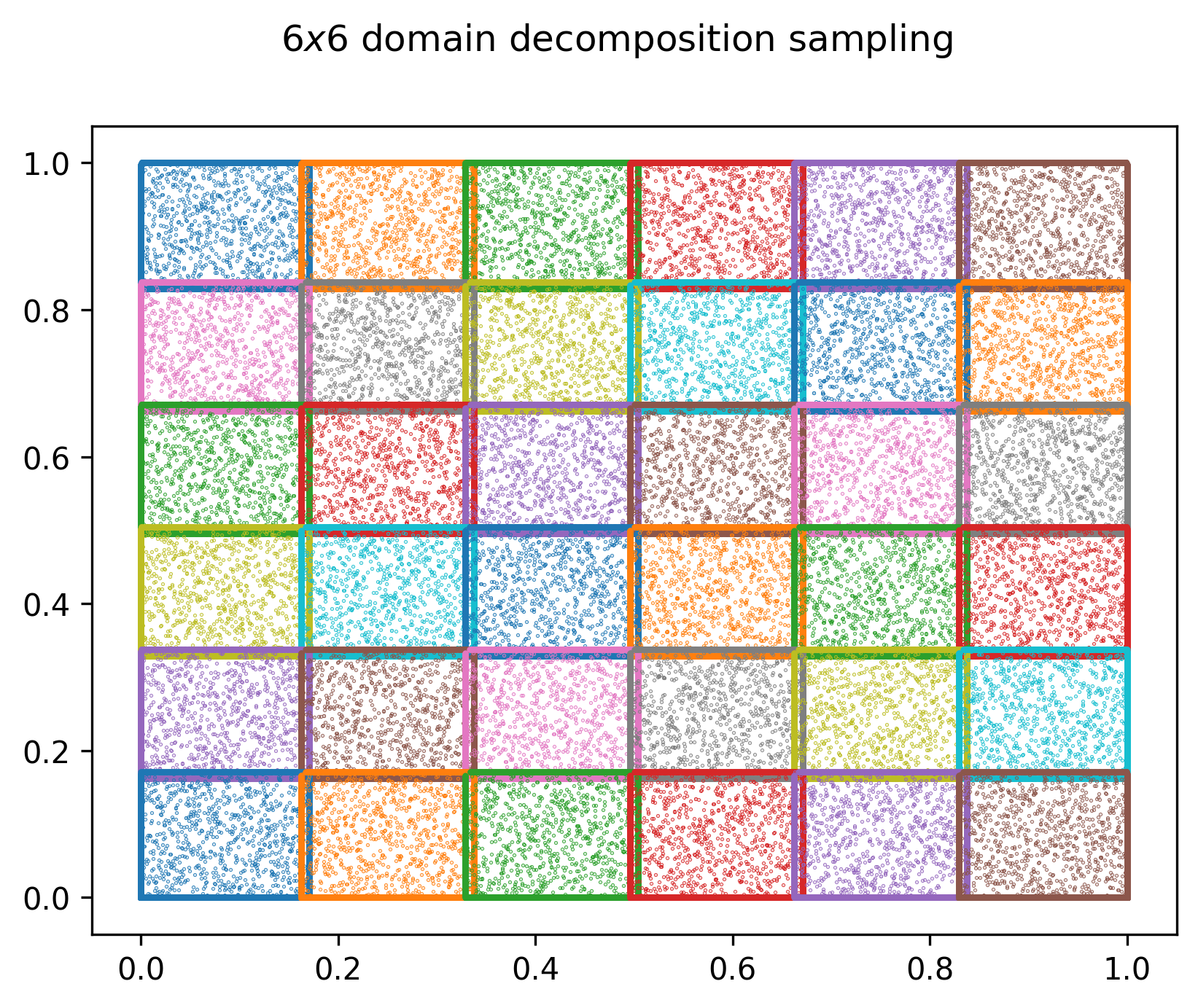}
\caption{(1) Strong scalability test for test problem with $\omega_1=1,\omega_2=3$ : Test with 2\,500 epoch (2) an example of sampling with a $6\times 6$ decomposition}
\label{fig:figure4}
\end{figure}
\begin{figure}[t]
\centering
\includegraphics[width=.5\textwidth]{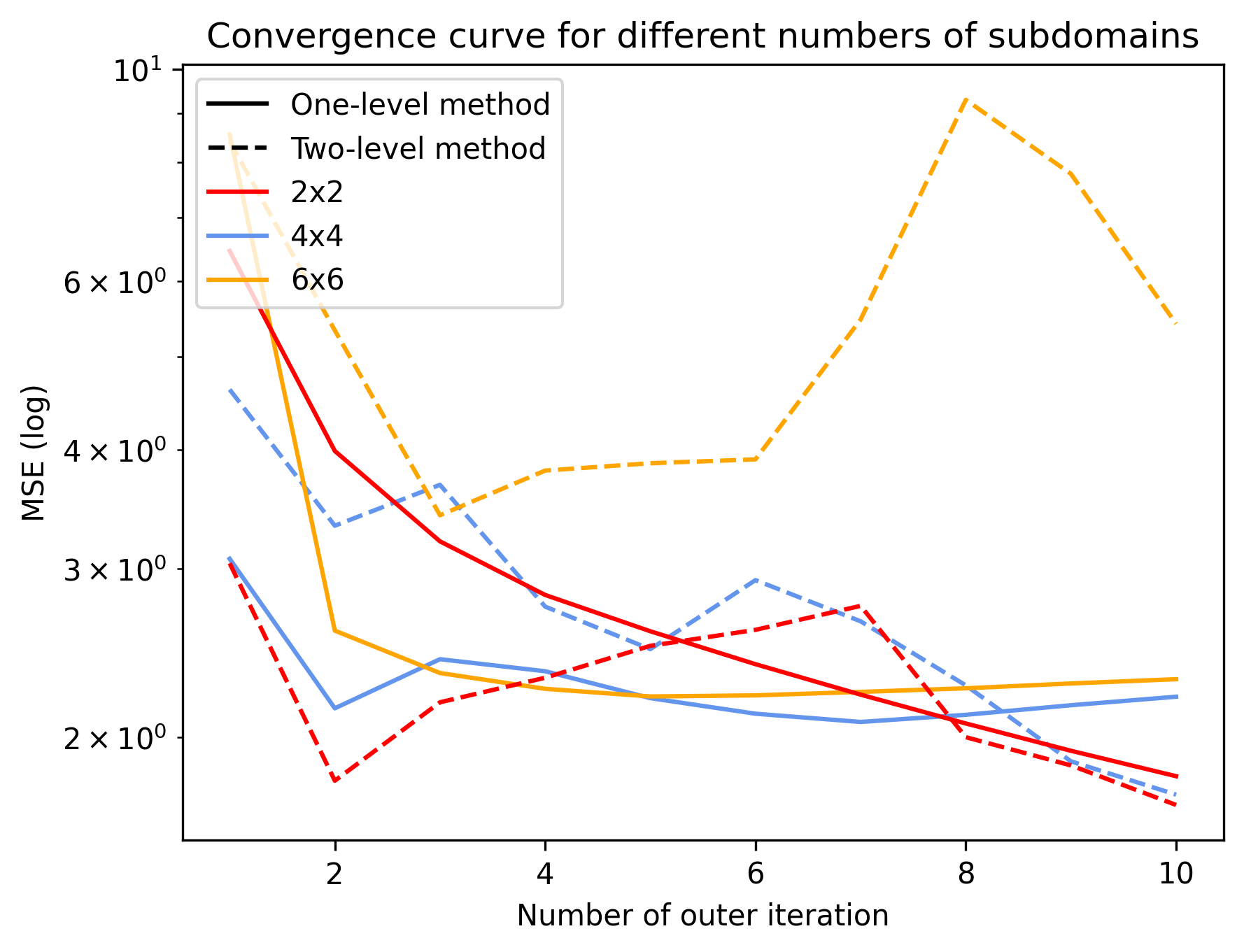}\hfill
\includegraphics[width=.5\textwidth]{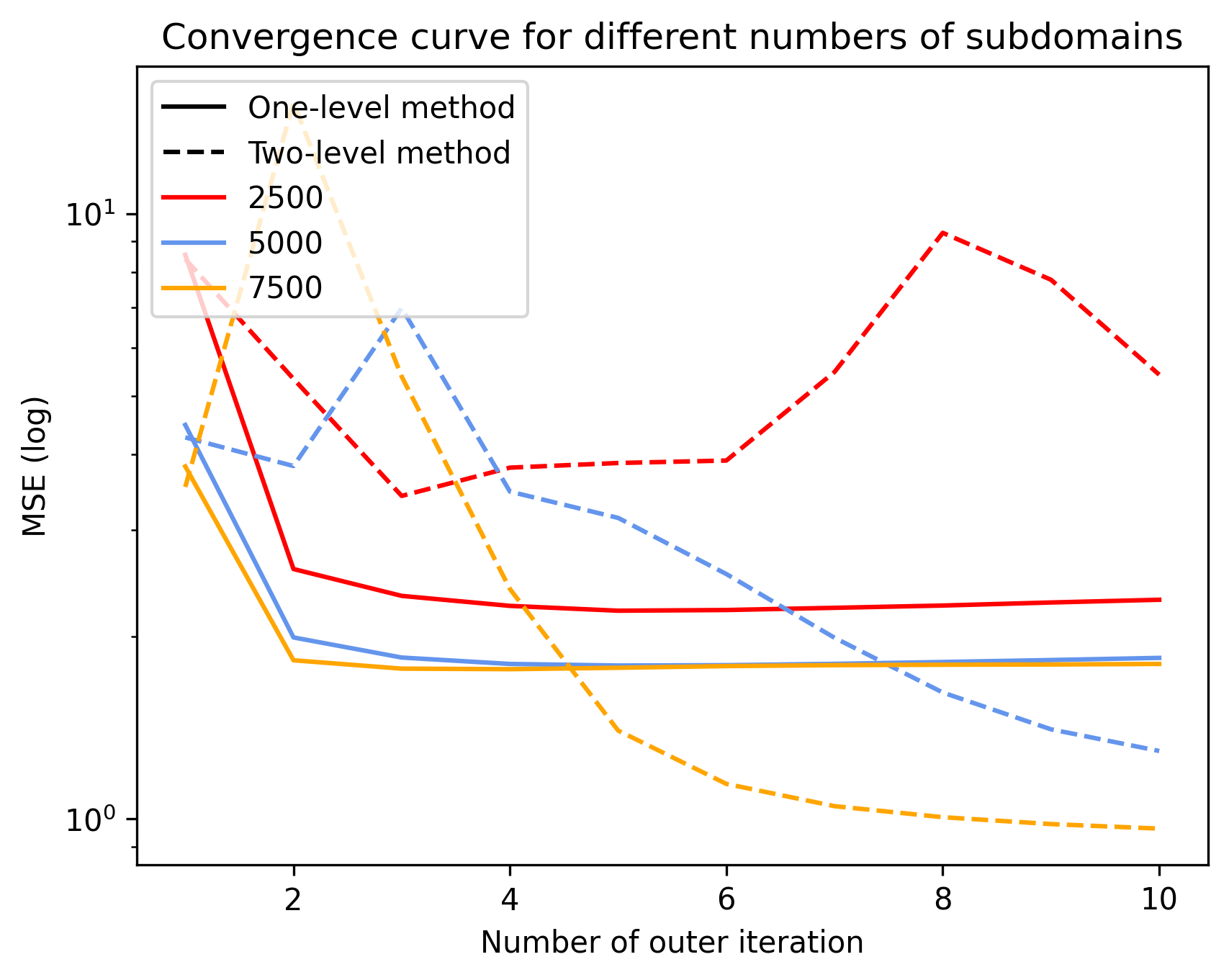}
\caption{Strong scalability test for test problem with $\omega_1=1,\omega_2=6$ : (1) Test with 2\,500 epoch (2) Test on $6\times 6$ decomposition with variation in the number of epochs}
\label{fig:figure5}
\end{figure}
For the lower frequency pair of coefficients, that is, $(\omega_1,\omega_2) = (1,3)$, we observe a noticeable improvement in convergence with the addition of a coarse level; cf.~\cref{fig:figure4}. In particular, the convergence seems to be independent of the number of subdomains. However, when we increase the frequency of the solution, that is, $(\omega_1,\omega_2) = (1,6)$, we notice that it takes longer for the two-level method before it converges; cf.~\cref{fig:figure5} (2). We observed that this problem can be easily solved by improving hyper parameter settings. In particular, for $6\times6$ subdomains, we observe that we can significantly improve the scalability by increasing the number of epochs for each subproblem; cf.~\cref{fig:figure5} (3). Notably, with $5\,000$ and $7\,500$ epochs per subproblem, the two-level methods clearly outperforms the one-level Deep-DDM. 
\begin{table}[t]
\centering
\begin{tabular}{|l|l|l|l|}
\hline
                 &  2500 epochs &5000 epochs &7500 epochs \\ \hline
One-level method & 66 min      & 148 min     &190 min     \\ \hline
Two-level method & 71 min      & 136 min    & 201 min     \\ \hline
\end{tabular}
\caption{Median of the wall time for the experiments of the figure \ref{fig:figure5} (2)}
\label{tab:walltime}
\end{table}

\section{Conclusion}

We have presented a two-level approach to improve the convergence of the one-level Deep-DDM method . The additional coarse networks facilitates faster global transport of information and enhances the scalability of the Deep-DDM method to larger numbers of subdomains. 
The cost of training the coarse network is relatively low compared with the total cost of the method and the method is well-suited for a parallel implementation with only small differences in wall time (\ref{tab:walltime} ) with a non optimized parallelization. 

\bibliography{ref}

\begin{thebibliography}{10}

\bibitem{dolean2023finite}
V.~Dolean, A.~Heinlein, S.~Mishra, and B.~Moseley.
\newblock Finite basis physics-informed neural networks as a {Schwarz} domain decomposition method, 2023.

\bibitem{dolean:MDD:2024}
V.~Dolean, A.~Heinlein, S.~Mishra, and Ben Moseley.
\newblock Multilevel domain decomposition-based architectures for physics-informed neural networks.
\newblock {\em Computer Methods in Applied Mechanics and Engineering}, 429:117116, 2024.

\bibitem{dolean_introduction_2015}
V.~Dolean, P.~Jolivet, and F.~Nataf.
\newblock {\em An {Introduction} to {Domain} {Decomposition} {Methods}}.
\newblock Society for Industrial and Applied Mathematics, 2015.

\bibitem{Gratton2024}
Serge Gratton, Valentin Mercier, Elisa Riccietti, and Philippe~L. Toint.
\newblock A block-coordinate approach of multi-level optimization with an application to physics-informed neural networks.
\newblock {\em Computational Optimization and Applications}, Aug 2024.

\bibitem{jang_partitioned_2023}
D.-K. Jang, K.~Kim, and H.~H. Kim.
\newblock Partitioned neural network approximation for partial differential equations enhanced with {Lagrange} multipliers and localized loss functions, December 2023.
\newblock arXiv:2312.14370 [physics].

\bibitem{klawonn2023machine}
A.~Klawonn, M.~Lanser, and J.~Weber.
\newblock Machine learning and domain decomposition methods -- a survey, 2023.

\bibitem{li_d3m_2020}
K.~Li, K.~Tang, T.~Wu, and Q.~Liao.
\newblock {D3M}: {A} deep domain decomposition method for partial differential equations.
\newblock {\em IEEE Access}, 8:5283--5294, 2020.
\newblock arXiv: 1909.12236.

\bibitem{li_deep_2020}
W.~Li, X.~Xiang, and Y.~Xu.
\newblock Deep {Domain} {Decomposition} {Method}: {Elliptic} {Problems}.
\newblock {\em arXiv:2004.04884 [cs, math]}, April 2020.
\newblock arXiv: 2004.04884.

\bibitem{moseley2021finite}
B.~Moseley, A.~Markham, and T.~Nissen-Meyer.
\newblock Finite basis physics-informed neural networks ({FBPINNs}): a scalable domain decomposition approach for solving differential equations, 2021.

\bibitem{raissi2017physics}
M.~Raissi, P.~Perdikaris, and G.~E. Karniadakis.
\newblock Physics informed deep learning (part i): Data-driven solutions of nonlinear partial differential equations, 2017.

\bibitem{schwarz_ueber_1870}
H.~A. Schwarz.
\newblock {\em Ueber einen {Grenzübergang} durch alternirendes {Verfahren}}.
\newblock Zürcher u. Furrer, 1870.

\bibitem{shukla_parallel_2021}
K.~Shukla, A.~D. Jagtap, and G.~E. Karniadakis.
\newblock Parallel {Physics}-{Informed} {Neural} {Networks} via {Domain} {Decomposition}.
\newblock {\em arXiv:2104.10013 [cs]}, April 2021.
\newblock arXiv: 2104.10013.

\bibitem{Wang_2021}
S.~Wang, H.~Wang, and P.~Perdikaris.
\newblock On the eigenvector bias of fourier feature networks: From regression to solving multi-scale pdes with physics-informed neural networks.
\newblock {\em Computer Methods in Applied Mechanics and Engineering}, 384:113938, October 2021.

\bibitem{Xu20}
Z.-Q.~J. Xu.
\newblock Frequency principle: Fourier analysis sheds light on deep neural networks.
\newblock {\em Communications in Computational Physics}, 28(5):1746--1767, 2020.

\end{thebibliography}
\bibliographystyle{plain}
\end{document}